\documentclass[letterpaper, 10 pt, conference]{ieeeconf}  % Comment this line out if you need a4paper

\IEEEoverridecommandlockouts                              % This command is only needed if 
\usepackage{graphics} % for pdf, bitmapped graphics files
\usepackage{epsfig} % for postscript graphics files
\usepackage{amsmath} % assumes amsmath package installed
\usepackage{amssymb}  % assumes amsmath package installed
\usepackage{array}
\usepackage{url}
\usepackage{verbatim}
\usepackage{graphicx}
\usepackage{cite}

\title{\LARGE \bf Hierarchical Reinforcement Learning for Articulated \\Tool Manipulation with Multifingered Hand
}

\author{Wei Xu,~\IEEEmembership{Student Member,~IEEE,} Yanchao Zhao,~\IEEEmembership{Student Member,~IEEE,} \\Weichao Guo,~\IEEEmembership{Member,~IEEE,} and Xinjun Sheng,~\IEEEmembership{Member,~IEEE}
\thanks{This work is supported in part by the Science and Technology Commission of Shanghai Municipality (Grant Nos. 24511103500, 23ZR1429700),  in part by the National Natural Science Foundation of China (Grant No. 52375021). (Corresponding authors: Xinjun Sheng; Weichao Guo)}
\thanks{The authors are with the State Key Laboratory of Mechanical System and Vibration, School of Mechanical Engineering, Shanghai Jiao Tong University, and Shanghai Key Laboratory of Intelligent Robotics, Meta Robotics Institute, Shanghai Jiao Tong University, Shanghai 200240, China. (e-mail: {\tt\footnotesize xu.wei@sjtu.edu.cn; zhaoyanchao1998@sjtu.edu.cn; guoweichao90@gmail.com; xjsheng@sjtu.edu.cn})}
\thanks{Accepted by IROS 2025. © 2025 IEEE. Final version to appear in IEEE Xplore.}
}

\begin{document}

\maketitle
\thispagestyle{empty}
\pagestyle{empty}

\begin{abstract}
    Manipulating articulated tools, such as tweezers or scissors, has rarely been explored in previous research. Unlike rigid tools, articulated tools change their shape dynamically, creating unique challenges for dexterous robotic hands. In this work, we present a hierarchical, goal-conditioned reinforcement learning (GCRL) framework to improve the manipulation capabilities of anthropomorphic robotic hands using articulated tools. Our framework comprises two policy layers: (1) a low-level policy that enables the dexterous hand to manipulate the tool into various configurations for objects of different sizes, and (2) a high-level policy that defines the tool's goal state and controls the robotic arm for object-picking tasks. We employ an encoder, trained on synthetic pointclouds, to estimate the tool's affordance states—specifically, how different tool configurations (e.g., tweezer opening angles) enable grasping of objects of varying sizes—from input point clouds, thereby enabling precise tool manipulation. We also utilize a privilege-informed heuristic policy to generate replay buffer, improving the training efficiency of the high-level policy. We validate our approach through real-world experiments, showing that the robot can effectively manipulate a tweezer-like tool to grasp objects of diverse shapes and sizes with a 70.8\% success rate. This study highlights the potential of RL to advance dexterous robotic manipulation of articulated tools. 

\end{abstract} 

\section{INTRODUCTION}
\begin{figure}[!ht]
    \centering
    \includegraphics[width=0.7\linewidth]{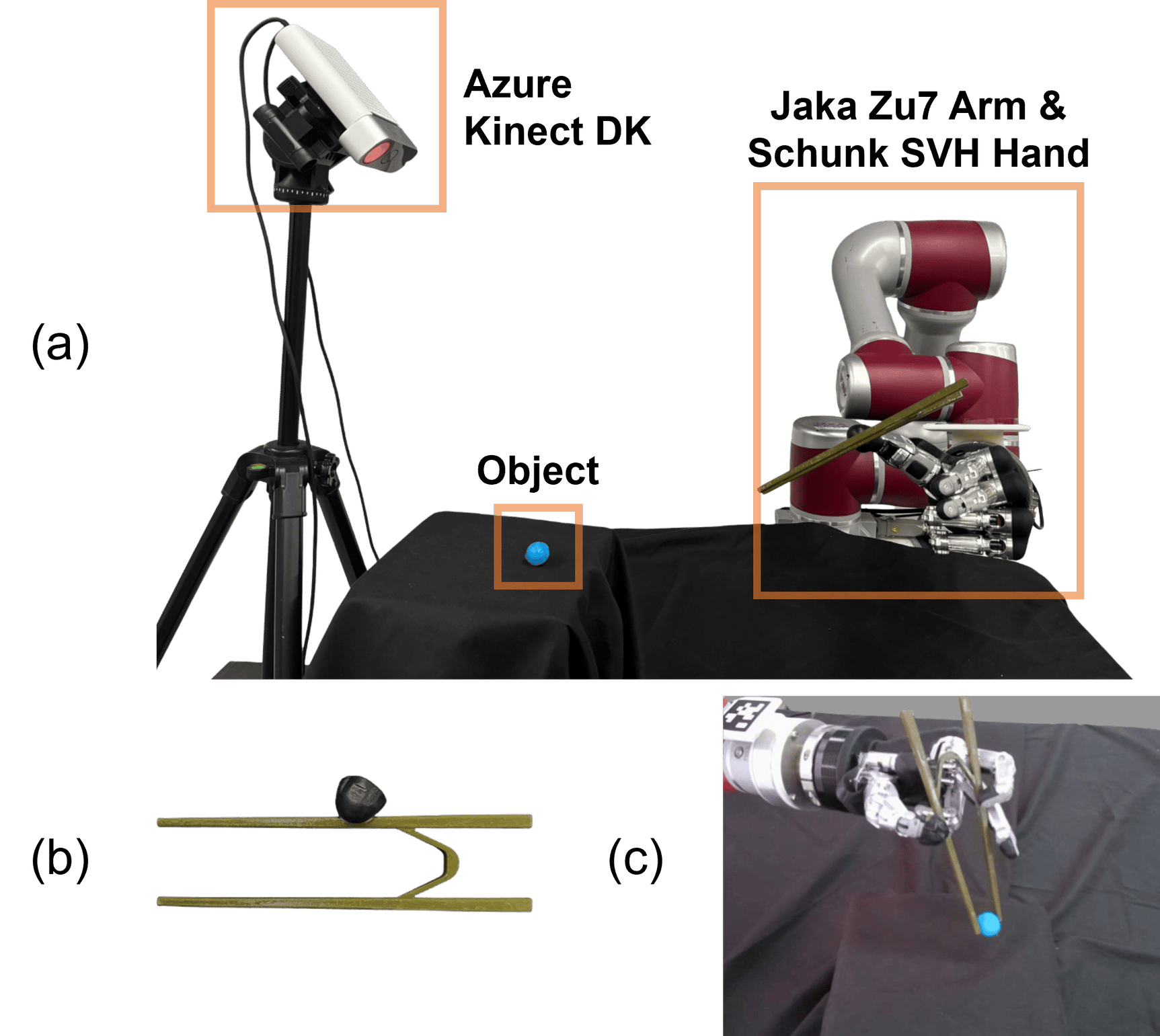}
    \caption{Experimental setup for tweezer-based manipulation. (a) The robotic platform includes an Azure Kinect DK camera, a Jaka Zu7 robotic arm, and a Schunk SVH hand. (b) A 3D-printed tweezer with a thumb slot for stability. (c) The task requires the robotic arm and dexterous hand to manipulate the tweezer and grasp the target object from the table.}
    \label{fig1}
\end{figure}

In recent years, significant progress has been made in the design of dexterous robotic hands, achieving unprecedented levels of dexterity \cite{Kim21Integrated}. These advancements have enabled numerous applications in social robotics and prosthetics \cite{Prasad22HumanRobot}. Existing studies extensively investigate dexterous hand grasping tasks, leading to higher efficiency and improved success rates \cite{Wei24Learning}. However, human hands can perform more than just grasping; they also enable complex in-hand manipulations, such as tool use—challenges not easily met by parallel grippers. Tool use is particularly challenging because the robotic hand must continuously adapt tool position and configuration, maintaining effective contact with environmental objects. 

Many researchers have applied RL to address the challenges of high-dimensional control and complex contact dynamics in dexterous grasping, achieving generalized capabilities for objects with intricate shapes \cite{Shaw24Learning, Hoang23Learning, Peri24Pointa, Pavlichenko23Deep, Liu23DexRepNet, Zhang23CherryPicking}. Moreover, significant efforts have explored advanced tasks such as in-hand reorientation, assembly, and valve turning \cite{Luo24Progressive, Zhu19Dexterous}. Recent work has extended RL to rigid tool use with dexterous hands, like hammers or brushes. Notably, prior research predominantly treats chopsticks as simple two-finger grippers, with limited exploration of multi-finger dexterous hands for tool manipulation and indirect object grasping \cite{Ze23HInDex, Xu23Dexterous}. However, everyday tools are not limited to fixed-shape tools. They also encompass a variety of articulated tools that perform diverse tasks by altering their geometry. Examples include chopsticks, tweezers, and scissors, which dynamically change shape to grasp or cut objects. Manipulating such articulated tools presents unique challenges: (1) Dynamic shape adaptation requires precise coordination of multiple fingers to control tool joint angles (e.g., tweezer opening); (2) Visual perception must simultaneously track tool pose, shape changes, and object position under occlusions; and (3) Hand-arm coordination demands different control strategies for gross arm movements versus fine finger adjustments. These challenges make articulated tool manipulation significantly more complex than rigid object grasping, leaving how to enable dexterous hands to adeptly manipulate these shape-changing tools an open and largely unexplored research question.

In this paper, we aim to enhance the ability of anthropomorphic hands to manipulate articulated tools, as illustrated in Fig. \ref{fig1}. Articulated tools, such as tweezers, play an important role in daily activities by frequently assisting human hands in grasping small objects within confined spaces. Current dexterous hands are limited by their degrees of freedom, often restricted to grasping larger objects and struggling to handle smaller ones. Articulated tools, like tweezers, can further enhance the dexterous hands' ability to grasp small objects. Specifically, we design a prototype scenario involving a robotic arm, a dexterous hand system, and a tweezer-like articulated tool. The goal is for the robot to learn how to use this tool to pick objects up from a table. We employ a goal-conditioned RL (GCRL) framework with two hierarchical policy layers to decompose a high-DoF system into manageable subproblems \cite{Liu22GoalConditioneda, Qian23GoalConditioned, Islam22Discrete, Nair18Visual}. The low-level policy coordinates the dexterous hand's individual fingers, altering the tool's shape to accommodate objects of various sizes. In parallel, the high-level policy provides target states for the tool, while also commanding the robotic arm to position both the hand and tool for object-grasping tasks. We synthesize the tool's point clouds, which encode its pose (rotation, translation) and shape states (e.g., opening angles for tweezers).	We employ single value decomposition (SVD) registration and train the encoder to estimate the tool's pose and affordance states, which characterize the tool's functional capabilities for grasping objects of different sizes based on its current shape configuration (e.g., tweezer opening width), from the input point cloud. Unlike directly estimating distances from raw point clouds, which is inherently noisy due to sensor limitations, partial occlusions, and calibration errors, our method mitigates these challenges. By encouraging the goal space coverage, we train the low-level policy to manipulate the tool toward diverse configurations for manipulating objects with various shapes. To train the high-level policy, we use a replay buffer generated by a privilege-informed heuristic controller, allowing the policy to learn how to dispatch arm commands and goals for object-picking tasks. We validate our approach in real-world experiments, showing that it successfully enables the robot to use the tool for picking objects of various shapes and sizes. 

We summarize our main contributions as follows.
\begin{itemize} 
    \item A hierarchical RL framework that separates tool-use tasks into a high-level policy for moving the tool and defining its goal shape, and a low-level policy for manipulating the tool according to the goal with a dexterous hand.
    \item We introduce a novel encoder to extract the tool's pose and affordance shape information. Using a latent space derived from synthetic tool pointclouds, we train a low-level tool manipulation policy for a dexterous hand.
    \item By employing a heuristic controller with privileged information to generate replay buffer, we boost the training efficiency of the high-level policy.
\end{itemize}

\section{RELATED WORK}
\subsection{Dexterous Hand Manipulation}
Classic analytical methods address dexterous manipulation by modeling the hand-tool system's kinematics and dynamics. They often assume full knowledge of contact points, forces, and friction properties \cite{Chavan-Dafle15Prehensile} while restricting contact to fingertips only\cite{Xu24Stochastic}. For instance, \cite{Shi17Dynamic} accomplishes sliding manipulation by modeling sliding dynamics and surface contacts at soft fingertips. However, acquiring manipulation data from visual or tactile sensors remains challenging. Predefined geometric rules cannot easily account for uncertainties in object shape, surface properties, or dynamic environmental perturbations. 

RL has shown remarkable success in in-hand manipulation with multi-fingered hands, often surpassing what traditional analytical methods can achieve \cite{Yu22Dexterous}. For instance, \cite{Andrychowicz20Learning} used RL to coordinate multi-finger movements for in-hand object reorientation using visual input. Similarly, \cite{Pavlichenko23Deep} introduced a multi-component reward function to achieve functional grasps for known object categories, leveraging both reorientation and repositioning. RL has also been applied to other dexterous hand tasks. For instance, \cite{Zhu19Dexterous} uses finger joint angles, valve angles, and door-opening angles to perform valve turning and door-opening tasks with a three-fingered robotic hand. Furthermore, \cite{Xu23Dexterous} enabled users to specify intermediate and final goals via images, training robots to accomplish human-defined goals without extensive reward engineering, including tasks like using a brush. Unlike prior dexterous manipulation tasks, which often involve rigid objects or tools for grasping and reorientation, we focus on a system comprising a robotic arm and a multifingered hand. In this setup, the dexterous hand manipulates a tweezer-like tool that changes its geometry, with each shape corresponding to a distinct affordance (e.g., different opening angles to grasp various object sizes). The robotic arm moves both the hand and tool to the target object and uses the tool to lift it.

\begin{figure*}[!ht]
    \centering
    \includegraphics[width=0.6\linewidth]{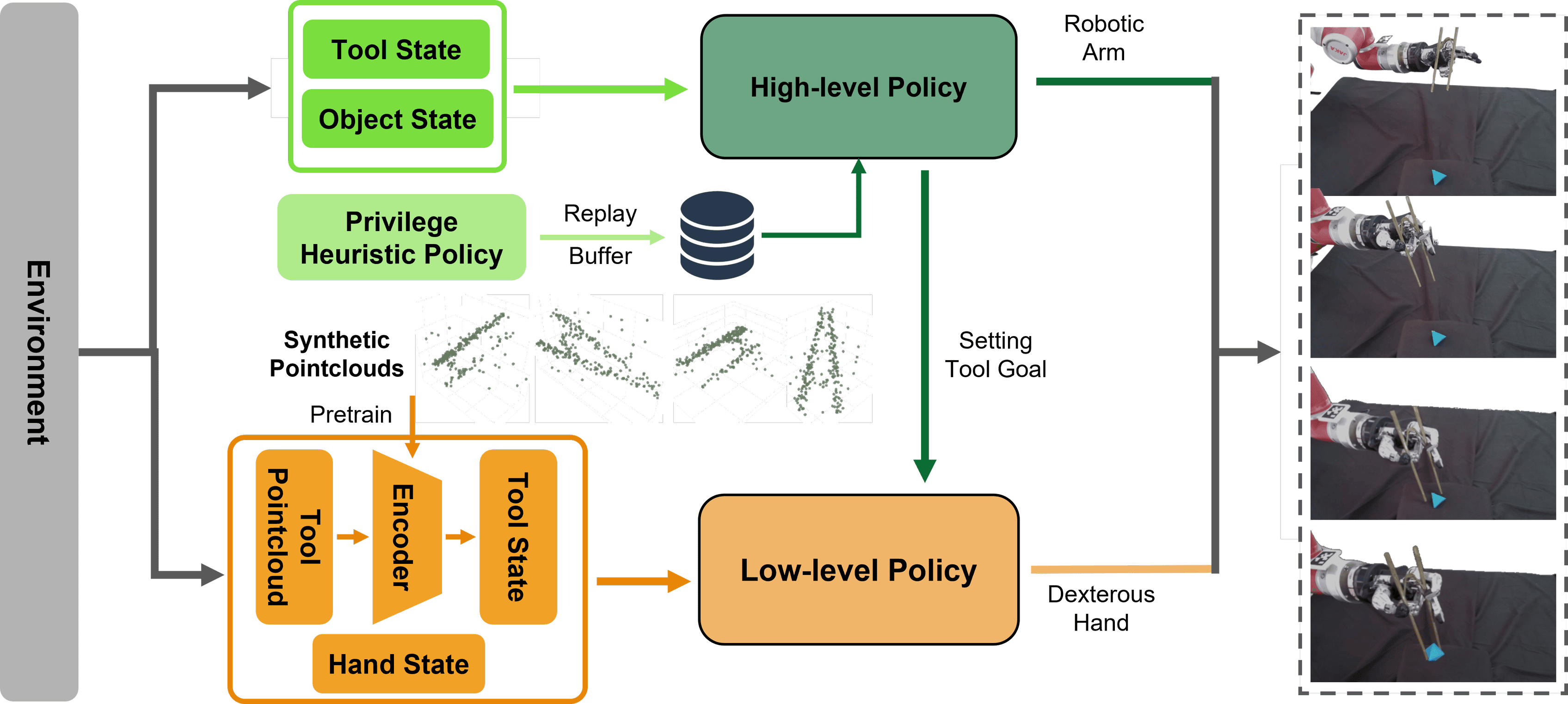}
    \caption{The proposed hierarchical framework for articulated tool manipulation. The low-level policy manipulates the tool with the dexterous hand, while the high-level policy provides tool-state goals and controls the robotic arm to position the tool.}
    \label{fig2}
\end{figure*}
\subsection{Object Representation in Dexterous Manipulation}
Robots require an object's geometry, pose, and affordances for dexterous manipulation. Previous approaches frequently assume that an object's position, pose, and velocity are known—simple in simulation but difficult in real-world settings \cite{Chen23Visual}. This gap complicates the deployment of simulation-trained policies in physical environments. Recent research addresses this issue by using RGB-D images to estimate object poses for manipulation tasks \cite{Andrychowicz20Learning, Morgan22Complex}. Additionally, \cite{Rana24AffordanceCentric} leveraged large vision models to detect object poses relative to affordance frames. Another research direction incorporates hand-object interaction as policy input. \cite{Liu23DexRepNet} introduced novel representations for hand-object interactions, capturing object surface features and the spatial relationship between hands and objects, allowing for generalized grasping. \cite{Chen23Visual} employed a teacher-student framework in which a teacher policy—given privileged information such as object pose and velocity—trained a student policy using only hand-object point clouds. This method successfully generalized in-hand object reorientation across diverse objects. 

In this study, merely obtaining the tool's pose and velocity is insufficient because tools like tweezers rely on deformation for their functionality. Consequently, extracting a tool's shape configuration from point clouds is crucial. We pre-train compact representations of tool shape, focusing specifically on tweezers. These representations encode how the tool shape is altered to perform tasks. Using GCRL, we train a low-level policy covering all possible tool shape configurations within the representation space. This approach enables a multi-fingered robotic hand to manipulate the tool effectively, adapting to objects of various shapes and sizes.

\section{METHODOLOGY}

\subsection{Problem Formulation}
In this work, we aim to enable dexterous multi-fingered hands to manipulate articulated tools for object-grasping tasks. Our proposed approach uses a GCRL framework (see Fig. \ref{fig2} for the method flowchart). Specifically, the low-level policy takes the tool state and hand state as inputs, manipulating the tool via the dexterous hand. The tool state is extracted through an encoder pre-trained on synthetic point cloud data, enabling precise representations of the tool's pose and affordances. We formulate the low-level problem as a Goal-Conditioned Markov Decision Process, defined as: $\mathcal{M}_G = (\mathcal{S}, \mathcal{A}, \mathcal{G}, \mathcal{T}, r, \gamma)$, where $ \mathcal{S}, \mathcal{A}, \mathcal{G}, \mathcal{T} $, and $ \gamma $ denote the state space, action space, goal space, transition dynamics, and discount factor, respectively. The goal-conditioned reward function $ r : \mathcal{S} \times \mathcal{A} \times \mathcal{G} \rightarrow \mathbb{R} $ is designed to encourage the low-level policy to achieve the designated goal efficiently and effectively. The high-level policy then takes the tool and object states as inputs, generates tool-state goals, and controls the robotic arm to position the tool. The low-level and high-level policies are trained separately in a sequential manner. First, we pre-train the low-level policy to master tool shape control using GCRL with synthetic point cloud data. After the low-level policy converges, we fix its parameters and train the high-level policy independently. Specifically, we employ a privilege-informed heuristic policy to generate a high-quality replay buffer containing hand-arm coordination trajectories, which significantly accelerates the high-level policy's training process by providing informative exploration guidance.

\subsection{Low-level Policy for Articulated Tool Manipulation}
To facilitate effective manipulation of geometry-altering tools with a multifingered hand, we employ a 3D-printed tweezer as a representative example. When using such tools, humans typically stabilize the tool's pose relative to their hand's coordinate frame while coordinating finger movements to adjust the tool's shape (e.g., changing tweezers' opening angles). Drawing inspiration from this behavior, we use RL to train a goal-conditioned low-level policy that enables the hand to precisely and adaptively control the tool's shape according to a specified goal state. 

\begin{figure}[!ht]
    \centering
    \includegraphics[width=0.8\linewidth]{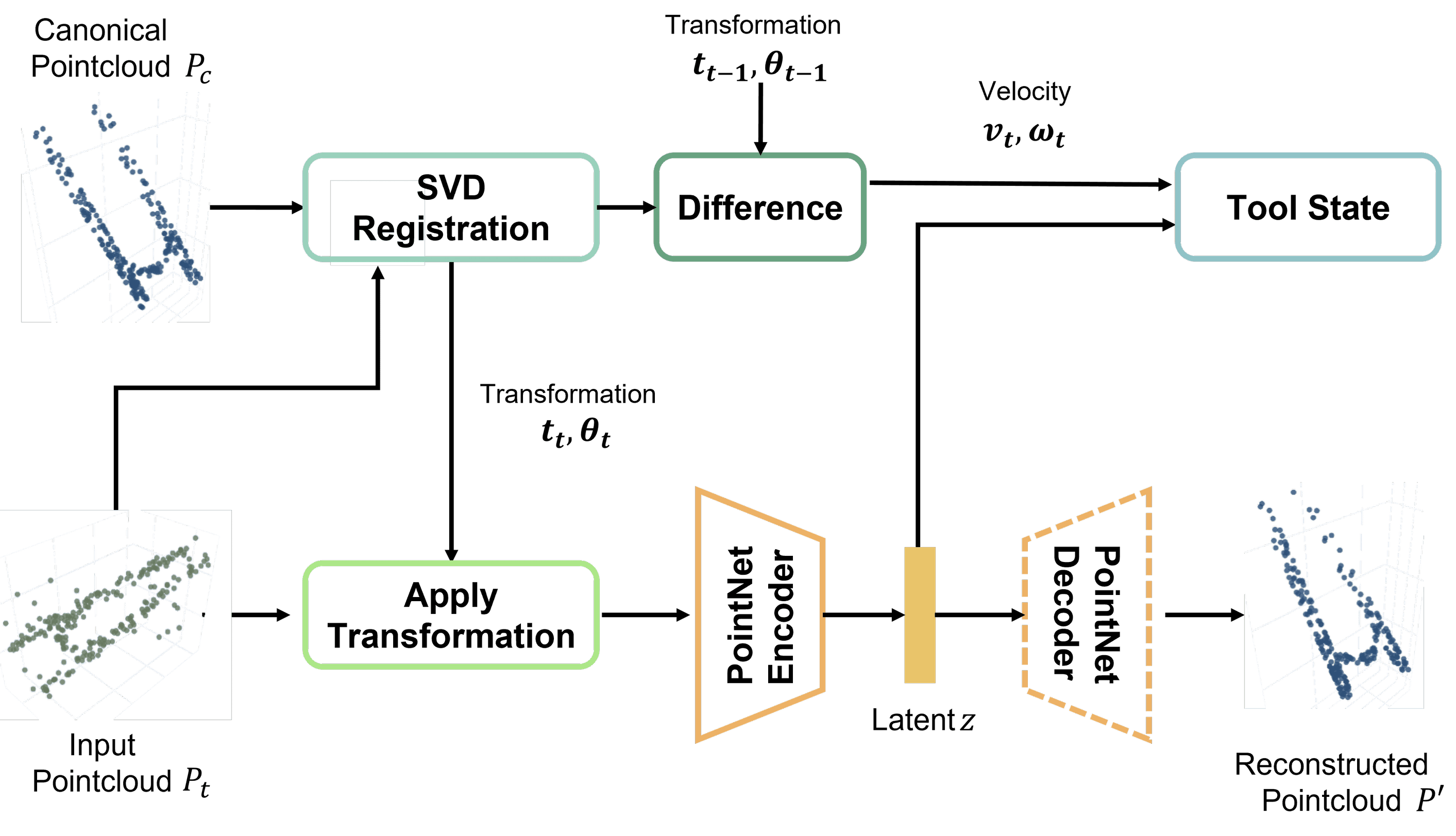}
    \caption{Architecture for extracting tool state from point clouds.}
    \label{fig3}
\end{figure}

A robust tool-state representation is crucial for effectively training and deploying the low-level policy. Fig. \ref{fig3} shows the pipeline for extracting the tool state in our approach. At each time step $ t $, we capture the tool's point cloud $ \mathbf P_t $ (e.g., tweezers) of 256 points. First, we apply SVD-based point cloud registration to compute coarse transformation parameters $ \mathbf{t}_t, \boldsymbol{\theta}_t $ between $\mathbf P_t $ and a canonical point cloud $\mathbf P_c $ \cite{Chen24Review, Bellekens14Survey}. Using the previous time step's transformation $ \mathbf{t}_{t-1}, \boldsymbol{\theta}_{t-1} $, we derive the tool's linear and angular velocities $ \mathbf{v}_t, \boldsymbol{\omega}_t $ (representing the global velocity of the tool frame relative to the hand's base frame, not individual tooltip velocities) via simple differencing. 

\subsubsection{Representation for Manipulating Articulated Tools}

Next, we transform $\mathbf P_t$ into the canonical coordinate frame, enabling a PointNet-based encoder to extract pose-invariant features primarily reflecting the tool's shape through shared multi-layer perceptrons and max pooling \cite{Charles17PointNet}. We then set the latent variable $\mathbf{z}$, representing tool shape including grasp aperture, to dimension 2. The encoder is pre-trained on a point cloud reconstruction task using 10k simulated input point clouds paired with corresponding ground truth data. To approximate real-world sensor data, we simulate occlusions by removing 20\% of the points, apply random rotations and translations to model tool-pose variations, and add Gaussian noise with a standard deviation of 5mm to each point. The encoder's loss function is $ \mathcal{L} = \mathcal{L}_{R} + \beta_t \mathcal{L}_{KL} $, where $ \mathcal{L}_{R} $ is a Chamfer-distance-based reconstruction loss, and $ \mathcal{L}_{KL} $ is a Kullback-Leibler divergence term that regularizes the latent distribution toward a standard Gaussian.

Finally, the tool representation comprises two components: (1) the tool's velocity and angular velocity $ \mathbf{v}_t, \boldsymbol{\omega}_t $ relative to the hand's coordinate frame, and (2) the latent representation $\mathbf{z}$, capturing the tool's affordances and shape. This compact representation offers the key information required for effective tool manipulation. 

\subsubsection{State and Action Space}
The state space of the low-level policy includes the multi-fingered hand state , and the tool state , and the given goal. The multifingered hand's state is given by current joint angles and velocities $\mathbf s_t^{hand}=[\mathbf q_t^{hand}, \dot{\mathbf q}_t^{hand}]$. The tool state $\mathbf s_t^{tool,low} = [\mathbf{v}_t, \boldsymbol{\omega}_t, \mathbf{z}_t]$ contains velocity $\mathbf{v}_t$, angular velocity $\boldsymbol{\omega}_t$ (defined in the hand's base frame), and the latent representation $\mathbf{z}_t$. To smooth the state and reduce high-frequency noise from sensor measurements and actuation delays, we apply an exponential moving average:
\begin{equation}
    \mathbf{\bar{s}}_t^{tool,low} = \alpha \mathbf s_t^{tool,low} + (1 - \alpha)\mathbf s_{t-1}^{tool,low}, \alpha = 0.9
\end{equation}

Finally, the low-level policy's state $\mathbf{s}_t^{low}$ is a 26-dimensional vector

\begin{equation}
    \mathbf s_t^{low} = [\mathbf s_t^{hand}, \mathbf {\bar{s}}_t^{tool,low}]
\end{equation}

We use a Schunk SVH hand with 20 joints and nine active degrees of freedom. The low-level policy runs at 20 Hz and outputs actions $\mathbf{a}_t^{low}$, which are the relative joint position changes. The target joint position for the hardware at the next time step is computed as $\mathbf{\hat{q}}_t^{hand} = \mathbf{q}_t^{hand} + \mathbf{a}_t^{low}$.

\subsubsection{Reward Design}
We introduce a dense reward to guide the policy toward the goal. This reward depends on the distance in the tool's latent space between the smoothed achieved goal $\bar{\mathbf{z}}_t$ and the desired goal $\mathbf{z}^{goal}$, encouraging the policy to minimize that distance. The desired goal $\mathbf{z}^{goal}$ is randomly sampled from $\mathcal{N}(0, 1)$. The goal reward is defined as:

\begin{equation}
    r^{goal}_t(\bar{\mathbf{z}}_t, \mathbf{z}^{goal}_t) = \exp(-c_{l1} ||\bar{\mathbf{z}}_t - \mathbf{z}^{goal}_t||_2)
\end{equation}
where $\bar{\mathbf{z}}_t$ is the smoothed latent variable for tool's shape, $c_{l1}> 0$ is a coefficient. Humans typically keep a tool's pose stable relative to the hand while minimizing finger movements. Inspired by this, we introduce an effort reward that penalizes unnecessary tool-pose changes and excessive finger movements, fostering stable and efficient manipulation. The effort reward is defined as

\begin{equation}
    \begin{split}
    r^{effort}_t(\bar{\mathbf{v}}_t, \bar{\boldsymbol{\omega}}_t, \dot{\mathbf{q}}_t^{hand}) = &-c_{l2} \sum_{\mathbf{p}_i \in \mathbf P_c} ||T(\bar{\mathbf{v}}_t, \bar{\boldsymbol{\omega}}_t)\mathbf{p}_i - \mathbf{p}_i||_2 \\
    & - c_{l3} ||\dot{\mathbf{q}}_t^{hand}||_2
    \end{split}
\end{equation}
where $ T(\bar{\mathbf{v}}_t, \bar{\boldsymbol{\omega}}_t) $ is the homogeneous transformation matrix defined by the tool's velocity $ \bar{\mathbf{v}}_t $ and angular velocity $ \bar{\boldsymbol{\omega}}_t $, and $c_{l2}, c_{l3}> 0$ are coefficients. We compute the Euclidean distance between the transformed and canonical point clouds to penalize large tool velocities.

The overall reward function for the low-level policy is defined as

\begin{equation}
    \begin{split}
    r^{low}_t(\mathbf{s}_t^{low}, \mathbf{a}_t^{low}, g^{low}) =& r^{goal}_t(\bar{\mathbf{z}}_t, \mathbf{z}^{goal}_t) + \\
    &r^{effort}_t(\bar{\mathbf{v}}_t, \bar{\boldsymbol{\omega}}_t, \dot{\mathbf{q}}_t^{hand})
    \end{split}
\end{equation}

With this reward function, the policy learns to control the hand's fingers to modify the tool's shape while maintaining stability, enabling efficient and precise manipulation.

\subsection{High-level Policy for Controlling Arm and Setting Goals}
After obtaining a robust low-level policy that can manipulate the tool in diverse configurations, we now introduce the high-level policy. The high-level policy sets goals for the low-level policy and controls the robotic arm, positioning the hand and tool near the target object. Together, the high-level and low-level policies enable the system to grasp and lift objects, completing the task successfully.

\subsubsection{State and Action Space}
The high-level policy's state space has two parts: the tool state and the object state. Unlike in the low-level policy, the high-level tool state is $ \mathbf{s}_t^{tool,high} = [\bar{\mathbf{t}}_t, \bar{\boldsymbol{\theta}}_t, \mathbf{z}_t] $, which includes the smoothed translation $ \mathbf{t}_t $, smoothed Euler angles $ \boldsymbol{\theta}_t $, and the latent representation $ \mathbf{z}_t $. 

The object state is $ \mathbf{s}_t^{obj} = [\mathbf{p}^{obj}_t, \mathbf{p}^{tgt}_t] $, where $\mathbf{p}^{obj}_t$ is the object's position and $\mathbf{p}^{tgt}_t$ is the target location for placing the object. Notably, the tool, object, and target positions are all defined in the hand's base frame, which is fixed to the robotic arm's end-effector. This ensures that the high-level policy depends solely on the relative positions of the tool and object, remaining invariant to different arm configurations. Consequently, the policy generalizes effectively across different object and target locations. Finally, the high-level policy's state $\mathbf{s}_t^{high}$ is a 14-dimensional vector

\begin{equation}
    \mathbf{s}_t^{high} = [\mathbf{s}_t^{tool,high}, \mathbf{s}_t^{obj}]
\end{equation}

We use a six-degree-of-freedom Jaka Zu7 robotic arm, with the high-level policy running at 20 Hz to match the low-level policy. The high-level policy outputs a five-dimensional vector $ \mathbf{a}_t^{high} = [\mathbf{z}^{goal}, \mathbf{a}_t^{arm}] $, where $ \mathbf{z}^{goal} $ is the low-level policy's goal state, and $ \mathbf{a}_t^{arm} $ is the velocity of the hand's coordinate frame relative to the arm's base. We compute the arm's joint velocities using the Jacobian pseudoinverse

\begin{equation}
    \dot{\boldsymbol{q}}^{arm} = \mathbf{J}^+ \mathbf{a}_t^{arm}, \mathbf{J}^+ = \mathbf{J}^T (\mathbf{J} \mathbf{J}^T)^{-1}
\end{equation}
where $\mathbf{J}^+$ is the pseudoinverse of the Jacobian matrix $ \mathbf{J}$. 

\subsubsection{Reward Design}
We define the high-level policy's reward function such that the task succeeds if the robot moves the target object to the specified goal, satisfying $||\mathbf{p}^{obj}_t - \mathbf{p}^{tgt}_t||_2 < 1\,\text{cm}$. Upon success, a sparse reward is given as

\begin{equation}
    r_{t}^{sparse} = c_{h1}, c_{h1}>0
\end{equation}

We also add a dense reward that encourages the tool to move closer to the target object and the object to approach the goal position.   The dense reward is

\begin{equation}
    \begin{split}
        r_{t}^{dense} = &\exp(-c_{h2} ||\mathbf{t}_t^{tool} - \mathbf{p}^{obj}_t||_2) + \\
        & c_{h3} \exp(-c_{h4} ||\mathbf{p}^{obj}_t - \mathbf{p}^{tgt}_t||_2)
    \end{split}
\end{equation}
where $c_{h2}, c_{h3}, c_{lh4} > 0$ are coefficients. For safety, we incorporate $r_{t}^{penalty} = -c_{h5}, c_{h5}>0$, penalizing cases where the distance between the tool and object or between the object and goal exceeds 20 cm, as well as any collisions with the table. The final high-level reward is

\begin{equation}
    r^{high}_t(\mathbf{s}_t^{high}, \mathbf{a}_t^{high}) = r^{sparse}_t + r^{dense}_t + r^{penalty}_t
\end{equation}

\subsubsection{Generating Replay Buffer with Privilege Heuristic Policy}
Due to the highly complex contact interactions involved in tool-based grasping—such as multi-point contacts between the dexterous hand and the tool, as well as contacts between the tool and the target object—training the high-level policy can be both data-intensive and time-consuming. Purely relying on random exploration often leads to slow convergence and high variance in performance. To address these challenges, we design a hand-crafted controller endowed with privileged states and actions to populate the replay buffer, thereby expediting the high-level policy's training.

In particular, our approach leverages privileged information unavailable to a standard learner, including precise spatial positions of the tweezer's two endpoints and direct control commands for the hinge joint. When the tweezer endpoints lie far from the target object, a simple proportional controller guides the tool toward the object, ensuring a robust approach trajectory. Once the endpoints are close to the target, we directly command the hinge joint to close the tweezers firmly around the object. As soon as the tweezer's inner sides match the object's diameter, indicating a stable grasp, the tool is moved to the goal position, thus completing the task. Throughout this process, we continuously record states for both the tool and the robotic arm, along with corresponding arm actions, and store them in the replay buffer. These data contain not only each time-step's precise tool state but also crucial information about hand-arm coordination. Moreover, we treat the current latent state of the tweezer as the high-level policy's action output from the previous time step, enabling the high-level policy to replay its decisions. By introducing replay buffer from our hand-crafted controller, the high-level policy rapidly learns how to manipulate the tweezer for object grasping, significantly reducing exploration time and improving training stability.

\subsection{Grasping Initialization and Reward Design}
Since stable tool grasping by the dexterous hand is critical, we clarify that the initial grasp is pre-defined with the tool firmly fixed in the hand. This setup eliminates grasping variability and focuses learning on manipulation rather than grasp acquisition.

For the reward function design, while explicit penalties for tool/object drops are not included, our experimental setup inherently minimizes these risks through the pre-fixed tool configuration. The reward primarily focuses on task success metrics (object pose, task completion) which implicitly encourage stable manipulation behaviors. This design choice prioritizes learning efficient manipulation strategies without over-constraining the policy with explicit penalty terms.
 
\section{EXPERIMENTS AND RESULTS}
\subsection{Simulated Experiments}
We use the MuJoCo simulator and the Soft Actor-Critic (SAC) algorithm to train both low-level and high-level policies \cite{Todorov12MuJoCo, Haarnoja18Soft}. Each policy network has two fully connected layers of 256 units each, with ReLU activations between layers. To bridge the sim-to-real gap, we implement various domain-randomization techniques.

For the low-level policy, we introduce variable state delays of up to 150 ms in the feedback loop, simulating the time misalignment often observed in real robotic and sensor systems.To reduce discrepancies between simulated and real robotic systems, we randomize control parameters using real hardware settings, adjusting them from 60\% to 150\% at the start of each episode. Additionally, we randomize friction coefficients for the fingers and tweezers from 0.2 to 1, assigning different coefficients to individual finger segments to reflect diverse materials (e.g., metal, rubber) used in real robotic hands. To enhance simulation efficiency, we synthesize the tool's point cloud by combining uniformly sampled points from its two components, applying their poses obtained from the simulator. To better approximate real conditions, we introduce random perturbations ($\pm 2$ cm in translation, $\pm30^{\circ}$ in Euler angles) to the pose of the tool pointcloud and add Gaussian noise (3 mm standard deviation) to each point. Additionally, we randomly remove 20\% of the points to simulate occlusions commonly encountered in real perception systems.

\begin{figure}[!ht]
    \centering
    \includegraphics[width=0.8\linewidth]{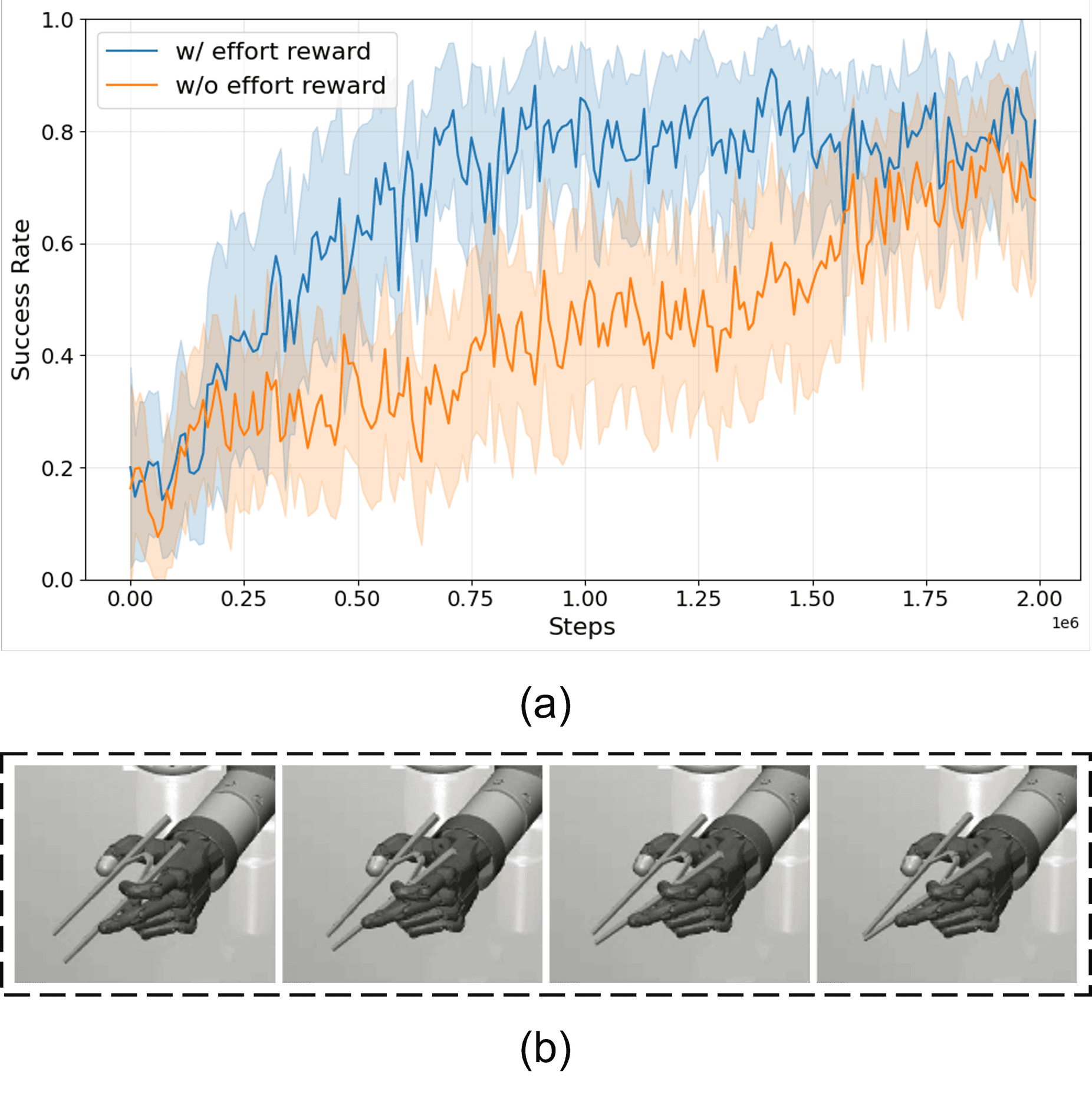}
    \caption{Training and evaluation of the low-level policy in simulation. (a) The training curve shows the success rates with and without the effort-based reward. (b) Visualization of the policy manipulating the tweezer to reach various tool-goal states.}
    \label{fig4}
\end{figure}

We conducted ablation studies in a simulated environment to examine how adding an effort reward affects training performance. Using three random seeds, we trained policies with and without effort-based rewards. As shown in Fig. \ref{fig4}(a), the results indicate that adding effort rewards significantly speeds up convergence and improves overall performance. The effort-based reward drives the policy to reduce unnecessary finger movements and keep the tool pose stable while manipulating its shape. This reward is critical for achieving stable and efficient training. Fig. \ref{fig4}(b) shows the low-level policy executing four randomly generated latent goals. The policy successfully controls the dexterous hand to manipulate the tweezers: the index finger hooks one side to stabilize it, while the thumb, middle, and index fingers coordinate to adjust the opening. This lets the tool accommodate objects of varying sizes. Although these manipulative motions differ from human use of tweezers or chopsticks—humans rely on soft fingertips and friction—such human-like strategies are less feasible for slippery, rigid robotic hands. Nonetheless, goal-conditioned RL with tool-shape representation effectively controls the dexterous hand, despite structural and frictional differences, allowing functionality similar to human manipulation.

\begin{figure}[!ht]
    \centering
    \includegraphics[width=0.8\linewidth]{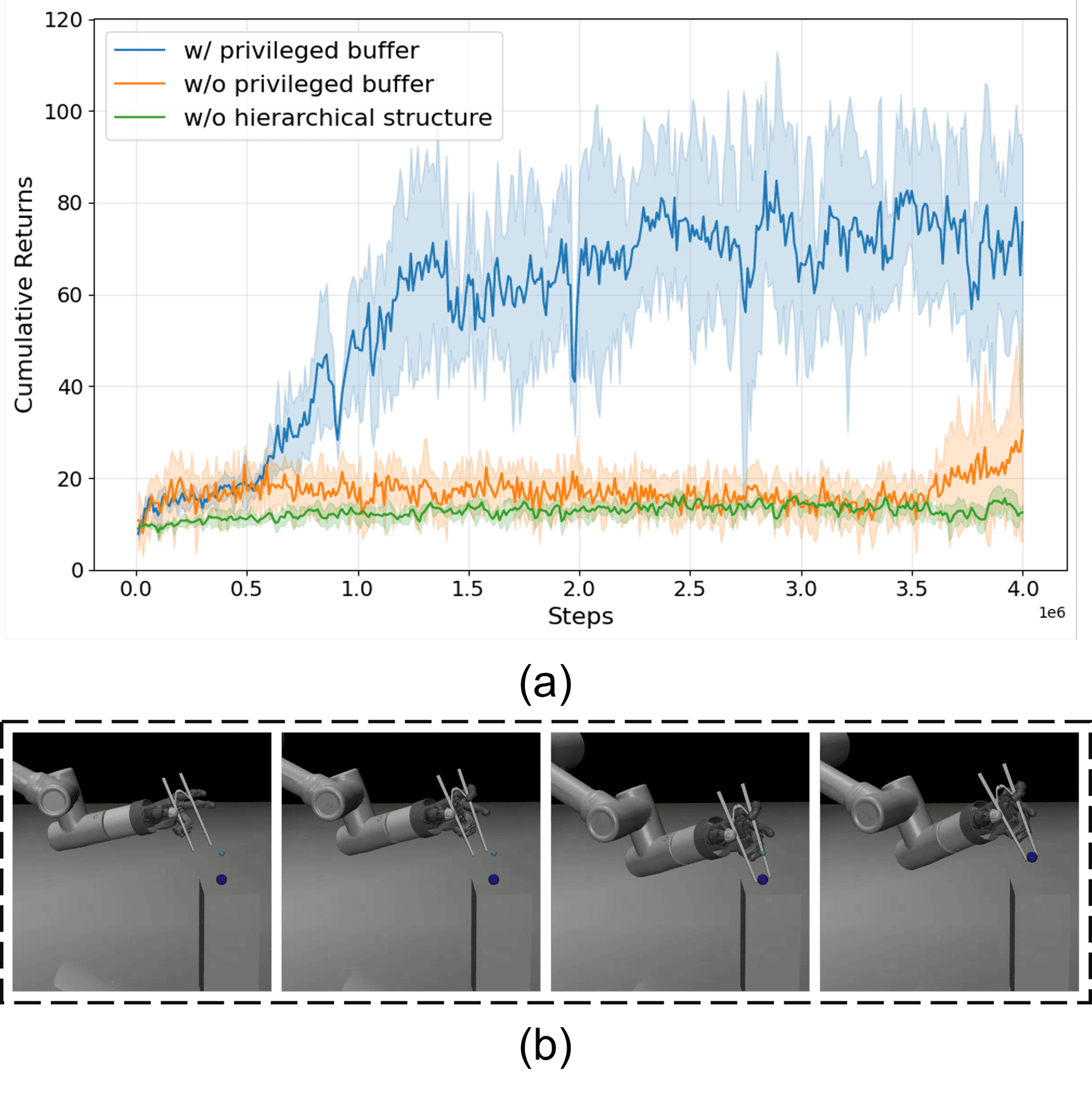}
    \caption{Training and evaluation of the high-level policy in simulation. (a) The training curve shows cumulative returns with and without the privileged buffer. (b) Sequential snapshots illustrate successful coordination of the arm and dexterous hand to grasp a small sphere.}
    \label{fig5}
\end{figure}

We applied domain-randomization to the high-level policy, introducing observation delays and randomizing arm-controller parameters. Within a $10\times10$ cm area, the target object—a dark blue sphere (1 cm radius, 10 g weight) shown in Fig. \ref{fig5}—was randomly placed, along with a final goal position 5 cm above it. We also randomized friction coefficients within the range of 1 to 5 to facilitate more stable training. To better approximate real conditions, we added Gaussian noise (2 mm standard deviation) to observed object positions, reflecting typical measurement inaccuracies in real robotic systems.

We conducted ablation studies in a simulated environment to examine the effect of a privilege-informed heuristic policy, used for replay buffer generation, on high-level policy training. Using three random seeds, we evaluated the policy under two conditions—one with the privileged buffer and one without. As shown in Fig. \ref{fig5}(a), introducing the privileged buffer significantly accelerates the training process. Moreover, by employing hierarchical RL rather than a non-hierarchical approach, we further improve learning efficiency. The privilege-informed heuristic policy supplies an imperfect yet highly informative replay buffer, markedly enhancing training stability. Fig. \ref{fig5}(b) illustrates the high-level policy successfully executing a tool-based object manipulation task. First, the robotic arm positions the tool near the target object (a dark blue sphere). The high-level policy then sends the desired tool configuration to the low-level policy, enabling a stable grasp. Lastly, the tool transports the object to a light blue sphere, representing the target location. These results confirm that the learned policy is capable of precise and effective tool-based manipulation.

\subsection{Real-World Validation}
In our real-world experiments, we used an Azure Kinect DK camera. First, we segmented the target object and tool from RGB images, then extracted their corresponding point clouds. For the tool's point cloud, we removed outliers, performed voxel-grid downsampling at 2 mm resolution, and randomly selected 256 points to create the input point cloud. For the object's point cloud, we calculated its centroid to determine the target object's position.  We then set the goal position 5 cm above the target object's centroid.

\begin{figure*}[!ht]
    \centering
    \includegraphics[width=0.8\linewidth]{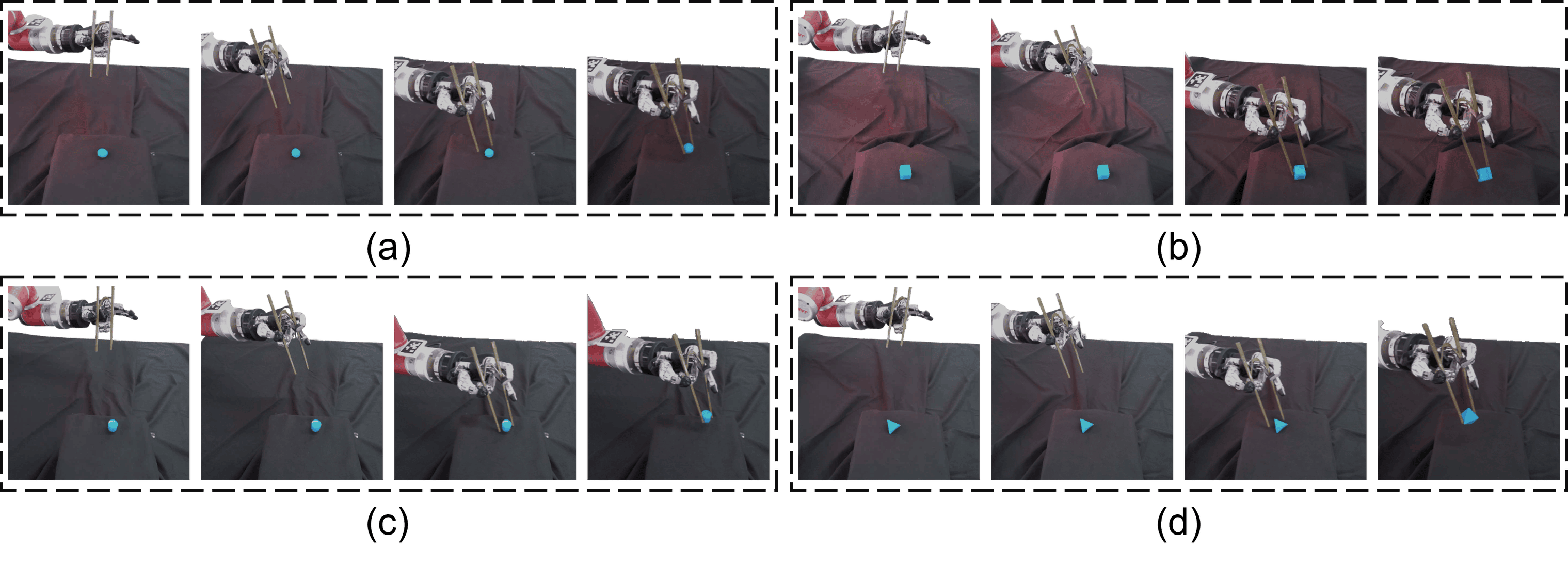}
    \caption{Demonstration of our method grasping various shapes with the tweezer: (a) Sphere, (b) Cube, (c) Cylinder, (d) Tetrahedron.}
    \label{fig8}
\end{figure*}

\begin{figure}[!h]
    \centering
    \includegraphics[width=0.8\linewidth]{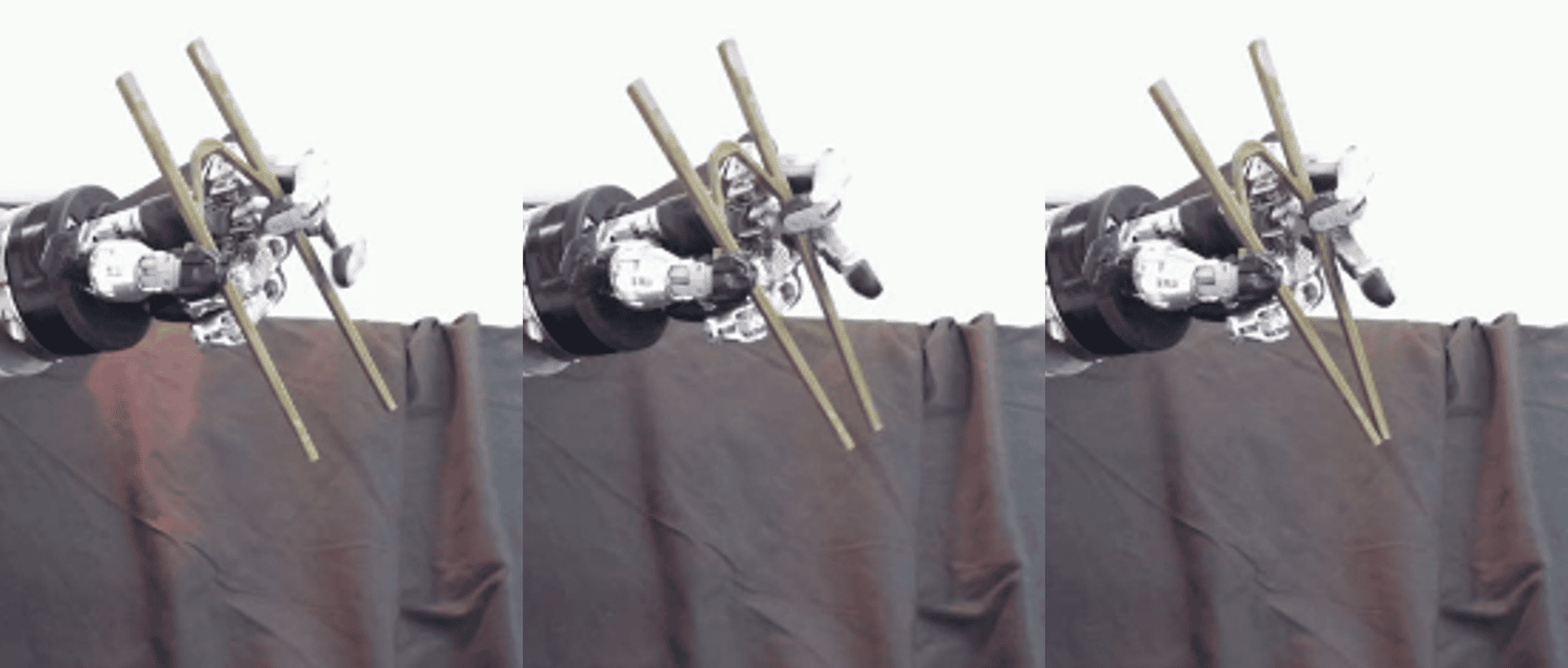}
    \caption{Demonstration of the low-level policy manipulating the tweezer into different configurations.}
    \label{fig6}
\end{figure}

We validated the low-level policy on a physical robot system. As shown in Fig. \ref{fig6}, the policy precisely controls the tweezers' shape with the robotic hand, demonstrating robustness and flexibility in manipulating articulated tools. 

We also validated the full hierarchical policy on a physical robot, using four objects of different shapes (sphere, cube, cylinder, tetrahedron) with dimensions from 1.5 to 2.5 cm. Each object was placed on a platform 20 cm high to avoid potentially dangerous collisions between the robotic arm or hand and the table. We performed 12 grasping trials per object. A trial was successful if the policy grasped the target object from the platform within 200 steps (about 10 seconds).

\begin{figure}[!h]
    \centering
    \includegraphics[width=0.7\linewidth]{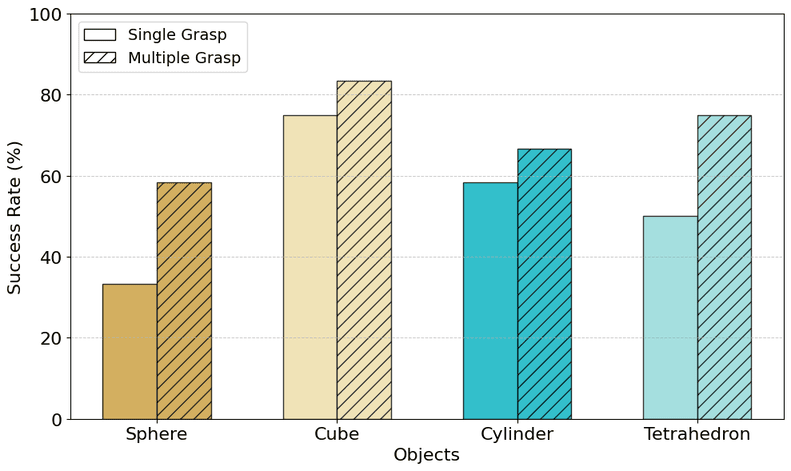}
    \caption{Success rates for object-picking tasks with four shapes: sphere, cube, cylinder, and tetrahedron.}
    \label{fig7}
\end{figure}

Fig. \ref{fig7} shows the success rates for the first attempt as well as the overall success rates after multiple attempts. The results indicate that the proposed policy can adaptively manipulate and grasp objects with diverse geometries, achieving an average success rate of 54.2\% on the first attempt. With multiple attempts, the success rate further improves to 70.8\%.

\begin{figure}[!h]
    \centering
    \includegraphics[width=0.7\linewidth]{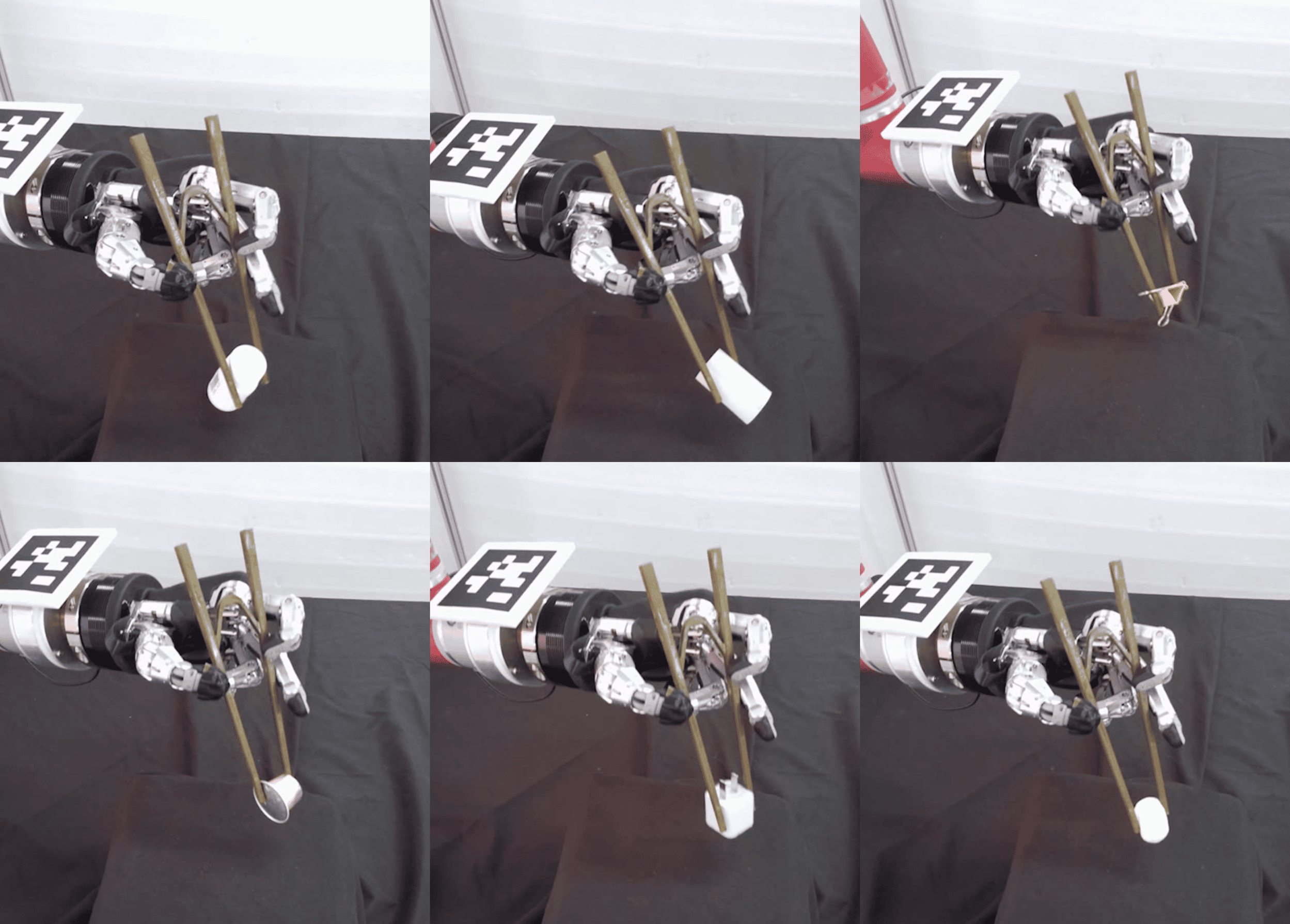}
    \caption{Successful grasps generated by our approach on six daily objects. The objects from top to bottom are vitamin bottle, tape, binder clip, coffee capsule, charger, and foam padding.}
    \label{fig10}
\end{figure}

Fig. \ref{fig8} shows the robotic system successfully grasping the four objects under the learned policy, highlighting the method's versatility and robustness for different shapes. In addition, we validate our approach in six daily objects, as shown in Fig. \ref{fig10}. While we do not explicitly incorporate target object geometry as input, the low-level policy's ability to generate diverse tool configurations allows the system to adapt to different object geometries.

\begin{figure}[!ht]
    \centering
    \includegraphics[width=0.8\linewidth]{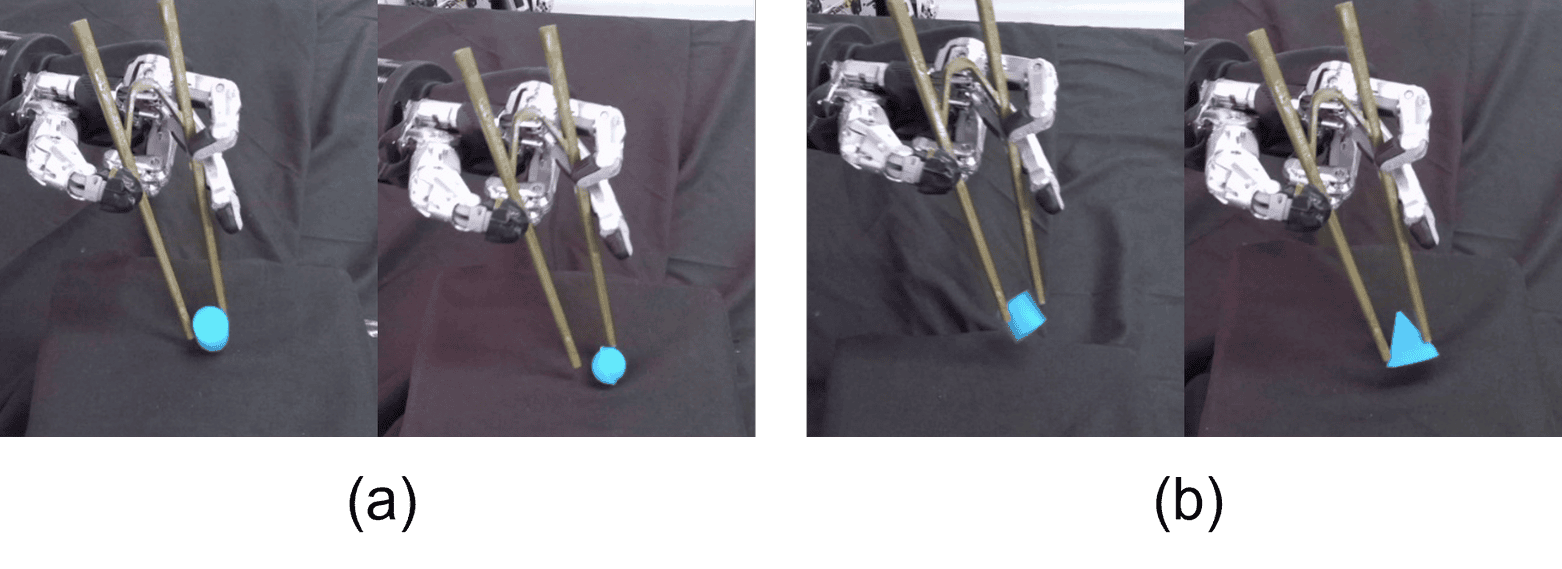}
    \caption{Failed grasping scenarios. (a) Inaccurate object position estimation. (b) Insufficient gripping force causing object slippage.}
    \label{fig9}
\end{figure}

We observed that failed grasps often arose from inaccurate object position estimates, frequently causing the first attempt to fail, as illustrated in Fig. \ref{fig9}(a). Interestingly, these failed attempts often shifted the object's pose, making it easier to grasp on subsequent tries and significantly boosting the success rate. This effect was particularly evident when grasping tetrahedral objects, as their orientation can drastically affect the difficulty of grasping. Other causes of failed grasps included insufficient grip force, leading to slippage, and lifting the object too early before achieving a secure grasp, as shown in Fig. \ref{fig9}(b). These issues may be attributed to the absence of fingertip tactile sensors in our robotic system. Without tactile feedback, the system relies solely on visual data, which makes it challenging to dynamically adjust its actions during grasping. In contrast, human hands utilize fine-grained control to prevent slippage and achieve stable grasps.

\section{DISCUSSION AND FUTURE WORK}
In this paper, we explored a novel dexterous hand manipulation task, in which the robotic hand reshapes an articulated hinge tool with its fingers and collaborates with a robotic arm for tool-based object grasping. To address this challenge, we proposed a hierarchical framework based on goal-conditioned RL. The high-level policy governs the robotic arm and sets tool-state goals, while the low-level policy manipulates the hinge tool via the dexterous hand. We employed SVD-based point cloud registration and a PointNet-based encoder to separate the tool's pose from its shape changes, forming a robust representation. Additionally, we added an effort reward and a privilege-informed heuristic policy for replay-buffer generation, significantly enhancing training efficiency. Experiments confirm the effectiveness of our representation and framework. Our method reached a 70.8\% success rate, demonstrating precise, adaptive tool manipulation. In summary, our study offers a promising solution for dexterous manipulation of articulated, deformable tools.

Despite these advances, our work still has limitations that should be addressed in future research. First, our study focuses on a single tool; future research could expand this approach to a wider array of hinge tools. Second, we assume the tool is already in the hand. An promising future direction is to integrate our framework with existing dexterous hand-grasping techniques, allowing the system to first learn tool grasping and then use the low-level policy for shape manipulation. This integration would form a complete pipeline from tool pickup to task execution.
\bibliographystyle{IEEEtran}
\bibliography{IEEEabrv, root.bib}

\begin{thebibliography}{10}
\providecommand{\url}[1]{#1}
\csname url@samestyle\endcsname
\providecommand{\newblock}{\relax}
\providecommand{\bibinfo}[2]{#2}
\providecommand{\BIBentrySTDinterwordspacing}{\spaceskip=0pt\relax}
\providecommand{\BIBentryALTinterwordstretchfactor}{4}
\providecommand{\BIBentryALTinterwordspacing}{\spaceskip=\fontdimen2\font plus
\BIBentryALTinterwordstretchfactor\fontdimen3\font minus \fontdimen4\font\relax}
\providecommand{\BIBforeignlanguage}[2]{{%
\expandafter\ifx\csname l@#1\endcsname\relax
\typeout{** WARNING: IEEEtran.bst: No hyphenation pattern has been}%
\typeout{** loaded for the language `#1'. Using the pattern for}%
\typeout{** the default language instead.}%
\else
\language=\csname l@#1\endcsname
\fi
#2}}
\providecommand{\BIBdecl}{\relax}
\BIBdecl

\bibitem{Kim21Integrated}
U.~Kim, D.~Jung, H.~Jeong, J.~Park, H.-M. Jung, J.~Cheong, H.~R. Choi, H.~Do, and C.~Park, ``Integrated linkage-driven dexterous anthropomorphic robotic hand,'' \emph{Nat Commun}, vol.~12, no.~1, p. 7177, Dec. 2021.

\bibitem{Prasad22HumanRobot}
V.~Prasad, R.~{Stock-Homburg}, and J.~Peters, ``Human-{{Robot Handshaking}}: {{A Review}},'' \emph{Int J of Soc Robotics}, vol.~14, no.~1, pp. 277--293, Jan. 2022.

\bibitem{Wei24Learning}
W.~Wei, P.~Wang, S.~Wang, Y.~Luo, W.~Li, D.~Li, Y.~Huang, and H.~Duan, ``Learning {{Human-Like Functional Grasping}} for {{Multifinger Hands From Few Demonstrations}},'' \emph{IEEE Trans. Robot.}, vol.~40, pp. 3897--3916, 2024.

\bibitem{Shaw24Learning}
K.~Shaw, S.~Bahl, A.~Sivakumar, A.~Kannan, and D.~Pathak, ``Learning dexterity from human hand motion in internet videos,'' \emph{The International Journal of Robotics Research}, vol.~43, no.~4, pp. 513--532, Apr. 2024.

\bibitem{Hoang23Learning}
M.-H. Hoang, L.~Dinh, and H.~Nguyen, ``Learning from {{Pixels}} with {{Expert Observations}},'' in \emph{2023 {{IEEERSJ Int}}. {{Conf}}. {{Intell}}. {{Robots Syst}}. {{IROS}}}, Oct. 2023, pp. 1200--1206.

\bibitem{Peri24Pointa}
S.~Peri, I.~Lee, C.~Kim, L.~Fuxin, T.~Hermans, and S.~Lee, ``Point {{Cloud Models Improve Visual Robustness}} in {{Robotic Learners}},'' in \emph{2024 {{IEEE Int}}. {{Conf}}. {{Robot}}. {{Autom}}. {{ICRA}}}, May 2024, pp. 17\,529--17\,536.

\bibitem{Pavlichenko23Deep}
D.~Pavlichenko and S.~Behnke, ``Deep {{Reinforcement Learning}} of {{Dexterous Pre-Grasp Manipulation}} for {{Human-Like Functional Categorical Grasping}},'' in \emph{2023 {{IEEE}} 19th {{Int}}. {{Conf}}. {{Autom}}. {{Sci}}. {{Eng}}. {{CASE}}}, Aug. 2023, pp. 1--8.

\bibitem{Liu23DexRepNet}
Q.~Liu, Y.~Cui, Q.~Ye, Z.~Sun, H.~Li, G.~Li, L.~Shao, and J.~Chen, ``{{DexRepNet}}: {{Learning Dexterous Robotic Grasping Network}} with {{Geometric}} and {{Spatial Hand-Object Representations}},'' in \emph{2023 {{IEEERSJ Int}}. {{Conf}}. {{Intell}}. {{Robots Syst}}. {{IROS}}}, Oct. 2023, pp. 3153--3160.

\bibitem{Zhang23CherryPicking}
Y.~Zhang, L.~Ke, A.~Deshpande, A.~Gupta, and S.~Srinivasa, ``Cherry-{{Picking}} with {{Reinforcement Learning}} : {{Robust Dynamic Grasping}} in {{Unstable Conditions}},'' \emph{ArXiv Prepr. ArXiv230305508}, Jun. 2023.

\bibitem{Luo24Progressive}
Y.~Luo, W.~Li, P.~Wang, H.~Duan, W.~Wei, and J.~Sun, ``Progressive {{Transfer Learning}} for {{Dexterous In-Hand Manipulation With Multifingered Anthropomorphic Hand}},'' \emph{IEEE Trans. Cogn. Dev. Syst.}, vol.~16, no.~6, pp. 2019--2031, Dec. 2024.

\bibitem{Zhu19Dexterous}
H.~Zhu, A.~Gupta, A.~Rajeswaran, S.~Levine, and V.~Kumar, ``Dexterous {{Manipulation}} with {{Deep Reinforcement Learning}}: {{Efficient}}, {{General}}, and {{Low-Cost}},'' in \emph{2019 {{Int}}. {{Conf}}. {{Robot}}. {{Autom}}. {{ICRA}}}, May 2019, pp. 3651--3657.

\bibitem{Ze23HInDex}
Y.~Ze, Y.~Liu, R.~Shi, J.~Qin, Z.~Yuan, J.~Wang, and H.~Xu, ``H-{{InDex}}: {{Visual Reinforcement Learning}} with {{Hand-Informed Representations}} for {{Dexterous Manipulation}},'' \emph{Adv. Neural Inf. Process. Syst.}, vol.~36, pp. 74\,394--74\,409, Dec. 2023.

\bibitem{Xu23Dexterous}
K.~Xu, Z.~Hu, R.~Doshi, A.~Rovinsky, V.~Kumar, A.~Gupta, and S.~Levine, ``Dexterous {{Manipulation}} from {{Images}}: {{Autonomous Real-World RL}} via {{Substep Guidance}},'' in \emph{2023 {{IEEE Int}}. {{Conf}}. {{Robot}}. {{Autom}}. {{ICRA}}}, May 2023, pp. 5938--5945.

\bibitem{Liu22GoalConditioneda}
M.~Liu, M.~Zhu, and W.~Zhang, ``Goal-{{Conditioned Reinforcement Learning}}: {{Problems}} and {{Solutions}},'' in \emph{Proc. {{Thirty-First Int}}. {{Jt}}. {{Conf}}. {{Artif}}. {{Intell}}.}\hskip 1em plus 0.5em minus 0.4em\relax Vienna, Austria: International Joint Conferences on Artificial Intelligence Organization, Jul. 2022, pp. 5502--5511.

\bibitem{Qian23GoalConditioned}
Z.~Qian, M.~You, H.~Zhou, X.~Xu, and B.~He, ``Goal-{{Conditioned Reinforcement Learning With Disentanglement-Based Reachability Planning}},'' \emph{IEEE Robot. Autom. Lett.}, vol.~8, no.~8, pp. 4721--4728, Aug. 2023.

\bibitem{Islam22Discrete}
R.~Islam, H.~Zang, A.~Goyal, A.~M. Lamb, K.~Kawaguchi, X.~Li, R.~Laroche, Y.~Bengio, and R.~{Tachet des Combes}, ``Discrete {{Compositional Representations}} as an {{Abstraction}} for {{Goal Conditioned Reinforcement Learning}},'' \emph{Adv. Neural Inf. Process. Syst.}, vol.~35, pp. 3885--3899, Dec. 2022.

\bibitem{Nair18Visual}
A.~V. Nair, V.~Pong, M.~Dalal, S.~Bahl, S.~Lin, and S.~Levine, ``Visual {{Reinforcement Learning}} with {{Imagined Goals}},'' \emph{Adv. Neural Inf. Process. Syst.}, vol.~31, 2018.

\bibitem{Chavan-Dafle15Prehensile}
N.~{Chavan-Dafle} and A.~Rodriguez, ``Prehensile pushing: {{In-hand}} manipulation with push-primitives,'' in \emph{2015 {{IEEERSJ Int}}. {{Conf}}. {{Intell}}. {{Robots Syst}}. {{IROS}}}, Sep. 2015, pp. 6215--6222.

\bibitem{Xu24Stochastic}
W.~Xu, Y.~Zhao, W.~Guo, and X.~Sheng, ``Stochastic {{Force-Closure Grasp Synthesis}} for {{Unknown Objects Using Proximity Perception}},'' \emph{IEEEASME Trans. Mechatron.}, pp. 1--11, 2024.

\bibitem{Shi17Dynamic}
J.~Shi, J.~Z. Woodruff, P.~B. Umbanhowar, and K.~M. Lynch, ``Dynamic {{In-Hand Sliding Manipulation}},'' \emph{IEEE Trans. Robot.}, vol.~33, no.~4, pp. 778--795, Aug. 2017.

\bibitem{Yu22Dexterous}
C.~Yu and P.~Wang, ``Dexterous {{Manipulation}} for {{Multi-Fingered Robotic Hands With Reinforcement Learning}}: {{A Review}},'' \emph{Front. Neurorobot.}, vol.~16, Apr. 2022.

\bibitem{Andrychowicz20Learning}
O.~M. Andrychowicz, B.~Baker, M.~Chociej, R.~J{\'o}zefowicz, B.~McGrew, J.~Pachocki, A.~Petron, M.~Plappert, G.~Powell, A.~Ray, J.~Schneider, S.~Sidor, J.~Tobin, P.~Welinder, L.~Weng, and W.~Zaremba, ``Learning dexterous in-hand manipulation,'' \emph{The International Journal of Robotics Research}, vol.~39, no.~1, pp. 3--20, Jan. 2020.

\bibitem{Chen23Visual}
T.~Chen, M.~Tippur, S.~Wu, V.~Kumar, E.~Adelson, and P.~Agrawal, ``Visual dexterity: {{In-hand}} reorientation of novel and complex object shapes,'' \emph{Sci. Robot.}, vol.~8, no.~84, p. eadc9244, Nov. 2023.

\bibitem{Morgan22Complex}
A.~S. Morgan, K.~Hang, B.~Wen, K.~Bekris, and A.~M. Dollar, ``Complex in-hand manipulation via compliance-enabled finger gaiting and multi-modal planning,'' \emph{IEEE Robot. Autom. Lett.}, vol.~7, no.~2, pp. 4821--4828, 2022.

\bibitem{Rana24AffordanceCentric}
K.~Rana, J.~{Abou-Chakra}, S.~Garg, R.~Lee, I.~Reid, and N.~Suenderhauf, ``Affordance-{{Centric Policy Learning}}: {{Sample Efficient}} and {{Generalisable Robot Policy Learning}} using {{Affordance-Centric Task Frames}},'' Oct. 2024.

\bibitem{Chen24Review}
L.~Chen, C.~Feng, Y.~Ma, Y.~Zhao, and C.~Wang, ``A review of rigid point cloud registration based on deep learning,'' \emph{Front. Neurorobotics}, vol.~17, p. 1281332, 2024.

\bibitem{Bellekens14Survey}
B.~Bellekens, V.~Spruyt, R.~Berkvens, and M.~Weyn, ``A survey of rigid 3d pointcloud registration algorithms,'' in \emph{{{AMBIENT}} 2014 {{Fourth Int}}. {{Conf}}. {{Ambient Comput}}. {{Appl}}. {{Serv}}. {{Technol}}. {{August}} 24-28 2014 {{Rome Italy}}}, 2014, pp. 8--13.

\bibitem{Charles17PointNet}
R.~Q. Charles, H.~Su, M.~Kaichun, and L.~J. Guibas, ``{{PointNet}}: {{Deep Learning}} on {{Point Sets}} for {{3D Classification}} and {{Segmentation}},'' in \emph{2017 {{IEEE Conf}}. {{Comput}}. {{Vis}}. {{Pattern Recognit}}. {{CVPR}}}.\hskip 1em plus 0.5em minus 0.4em\relax Honolulu, HI: IEEE, Jul. 2017, pp. 77--85.

\bibitem{Todorov12MuJoCo}
E.~Todorov, T.~Erez, and Y.~Tassa, ``{{MuJoCo}}: {{A}} physics engine for model-based control,'' in \emph{2012 {{IEEERSJ Int}}. {{Conf}}. {{Intell}}. {{Robots Syst}}.}, Oct. 2012, pp. 5026--5033.

\bibitem{Haarnoja18Soft}
T.~Haarnoja, A.~Zhou, P.~Abbeel, and S.~Levine, ``Soft {{Actor-Critic}}: {{Off-Policy Maximum Entropy Deep Reinforcement Learning}} with a {{Stochastic Actor}},'' in \emph{Proc. 35th {{Int}}. {{Conf}}. {{Mach}}. {{Learn}}.}\hskip 1em plus 0.5em minus 0.4em\relax PMLR, Jul. 2018, pp. 1861--1870.

\end{thebibliography}

\vfill

\end{document}